\documentclass[preprint,12pt,authoryear]{elsarticle}
\journal{Solar Energy}
\usepackage[utf8]{inputenc}
\usepackage{graphicx}
\usepackage{amssymb}
\usepackage{amsmath}
\usepackage{multirow}
\usepackage{booktabs,tabu}

\usepackage{microtype}

\usepackage[hidelinks]{hyperref}
\usepackage{natbib}
\let\cite\citep
\let\citebegin\citet

\usepackage{color}

\newcommand{\revise}[1]{#1}

\usepackage{fancyhdr}
\begin{document}
\begin{frontmatter}
\title{\vspace*{-1in}Deep Photovoltaic Nowcasting}
\author{Jinsong Zhang$^\dagger$, Rodrigo Verschae$^\mathsection$, Shohei Nobuhara$^\ddagger$, \\ and Jean-François Lalonde$^\dagger$\corref{cor1}}
\address{$^\dagger$Université Laval, 1065 ave de la médecine, Québec City, QC, G1V 0A6, Canada \\
$^\mathsection$Institute of Engineering Sciences, Universidad de O’Higgins, Rancagua, Chile \\
$^\ddagger$Kyoto University, Yoshidahonmachi, Sakyo, Kyoto 606-8501, Japan}

\cortext[cor1]{Corresponding author: \texttt{jflalonde@gel.ulaval.ca}. Part of this work was done while Rodrigo Verschae was at Kyoto University.}
\begin{abstract} 
Predicting the short-term power output of a photovoltaic panel is an important task for the efficient management of smart grids. Short-term forecasting at the minute scale, also known as \emph{nowcasting}, can benefit from sky images captured by regular cameras and installed close to the solar panel. However, estimating the weather conditions from these images---sun intensity, cloud appearance and movement, etc.---is a very challenging task that the community has yet to solve with traditional computer vision techniques. In this work, we propose to learn the relationship between sky appearance and the future photovoltaic power output using deep learning. We train several variants of convolutional neural networks which take historical photovoltaic power values and sky images as input and estimate photovoltaic power in a very short term future. In particular, we compare three different architectures based on: a multi-layer perceptron (MLP), a convolutional neural network (CNN), and a long short term memory (LSTM) module. We evaluate our approach quantitatively on a dataset of photovoltaic power values and corresponding images gathered in Kyoto, Japan. Our experiments reveal that the MLP network, already used similarly in previous work, achieves an RMSE skill score of 7\% over the commonly-used persistence baseline on the 1-minute future photovoltaic power prediction task. Our CNN-based network improves upon this with a 12\% skill score. In contrast, our LSTM-based model, which can learn the temporal dependencies in the data, achieves a 21\% RMSE skill score, thus outperforming all other approaches.
\end{abstract}

\begin{keyword}
short term forecast \sep deep learning \sep neural networks \sep computer vision
\end{keyword}
\end{frontmatter}

\section{Introduction}

While very attractive from environmental and economic perspectives, renewable energy sources such as wind and solar can provide significant challenges since energy production depends on external weather factors that cannot be controlled. For example, the power generated by a wind farm depends on the quantity of wind on a given day. Similarly, power generated by a solar panel depends on the cloud cover, sun position, weather conditions, etc. all of which can dramatically vary throughout the course of a single day. If solar power is to be fully developed, integrated into existing power grids and efficiently managed, the \emph{future} power output must be known.  

For this reason, the problem of forecasting future power output from solar panels has been studied extensively in the literature. Indeed, a wide variety of solutions for mid- to long-term forecasting have been proposed~\cite{olatomiwa2015support,bhardwaj2013estimation}. These techniques typically exploit weather stations (numerical weather predictions, or NWP) and/or satellite providing rich meta-data on which methods can rely. 

However, while mid- and long-term prediction are useful tasks, it is also important that we consider the problem of forecasting power output at the minute scale. This is known as short-term forecasting, or \emph{nowcasting}~\cite{lipperheide2015embedded}. Nowcasting is critical when managing operations of the smart grid, such as system integration, ensuring power continuity and managing ramp rates, etc. In this context, \revise{most} NWP and satellite data become unsuitable because of their low spatial and temporal resolution. Indeed, weather stations are \revise{typically} few and sparse, so the weather station closest to a solar panel might be too far to be reliable, and geostationary satellite have \revise{relatively} limited resolution. These limitations have been reported by several works~\cite{lipperheide2015embedded,rana2016univariate,russo2014genetic}. \revise{While some geostationary satellites (such as Himawari-8, etc.) do provide high temporal and spatial coverage and limited area models with high resolution NWP are also available, they still represent expensive options which might not always be easily accessible.}

A solution to this problem is to capture the local weather conditions at the solar panel at high spatial and temporal resolutions with a regular video camera pointing towards the sky and installed close to the panel~\cite{chow2011intra,marquez2013intra,urquhart2013sky,yang2014solar}. But while they are cheap and easy to install, they do not explicitly provide relevant weather information: rather, they provide images of the sky which must be analyzed in order to determine what is the relation between the images and the photovoltaic power output. What makes this analysis particularly challenging is the modeling of clouds. Their dynamics---including variations in shape, appearance, velocity, direction---create significant challenges to computer vision techniques which attempt to explicitly model and predict the future appearance of clouds. 

In this work, we bypass having to explicitly model cloud movement using traditional image processing techniques as was done in the previous work. Rather, we train algorithms that \emph{automatically learn} the relationship between the sky appearance (including clouds, sun, clear sky, etc.) and the photovoltaic power output of a solar panel. In particular, we rely on state-of-the-art deep learning techniques which efficiently learn to combine past power outputs and past images together into a compact model and accurately predict the future power output. Thus, our approach makes the explicit assumption that the current power output from the panel can be obtained. Of interest, we explore the use of several different deep learning techniques, including multi-layer perceptrons (MLP), convolutional neural networks (CNNs) and long short-term memory networks (LSTMs) applied to the problem of photovoltaic nowcasting. 

In short, we make the following three key contributions. First, \revise{we demonstrate that deep learning architectures can be used to predict the 1-minute future photovoltaic power output from past photovoltaic values and sky images captured with an off-the-shelf RGB camera, without the need for NWP or satellite data.} Second, we present three deep learning architectures adapted to that task. Finally, we present extensive experiments on a large and challenging dataset \revise{that evaluate and compare these architectures}. \revise{These experiments reveal that the deep learning models based on the LSTM structure outperform the widely-used baseline persistence model by an RMSE skill score of 21\% on the 1-minute photovoltaic power forecasting problem.} 

\revise{The rest of the paper is organized as follows. After reviewing relevant related work in sec.~\ref{section.related}, we briefly describe the data capture procedure in sec.~\ref{section.data} followed by an introduction to the notation used in the paper in sec.~\ref{section.overview}. Sec.~\ref{section.methods} then introduces the three different deep learning architectures used to predict photovoltaic production. Afterwards, experimental results are presented in sec.~\ref{section.experiments}, which is followed by a discussion on the limitations of our approach in sec.~\ref{section.discussion}. Finally, we conclude in sec.~\ref{section.conclusion} with promising future directions. }

\section{Related work}
\label{section.related}

This section briefly introduces approaches and data used for short-term photovoltaic forecasting, as well as relevant deep learning approaches. 

\paragraph{Approaches} The two main approaches for solar energy forecasting are dubbed the physical (or parametric) and data-driven methods. Physical models highly depend on numerical weather predictions (NWP), such as temperature and wind predictions, which are not typically suitable for very short term forecasting~\cite{antonanzas2016review,sobri2018solar}. On the contrary, data-driven methods have been widely used as ``black/gray-box'' models for short term solar energy forecasting by learning the relationship between historical data and solar energy production. A recent review~\cite{voyant2017machine} covers most machine learning methods used for solar energy forecasting. 

Among the different machine learning algorithms used for solar forecasting, Artificial Neural Networks (ANNs) have widely been used. In particular, numerous studies have reported that ANNs can provide more accurate results than physical models~\cite{hontoria2005application, jiang2008prediction, mellit201024,yadav2014solar}. 
A particular type of ANN, the Multilayer Perceptron (MLP) network has been used for solar energy forecasting, and has shown the ability to predict solar irradiance~\cite{mellit201024}. In our work, we compare more powerful deep learning architectures to the MLP model, and show these more recent techniques (namely, Convolutional Neural Networks or CNNs) do achieve higher performance on the short-term photovoltaic forecasting task. 

\paragraph{Data} Numerical weather predictions (NWP) and satellite images are widely used in solar energy forecasting since they provide valuable information such as cloud position and size, wind direction and speed for a large area. However, NWP and satellite images typically do not have adequate temporal resolution for very short term forecast~\cite{chow2011intra}. 

For very short term forecast,~\citebegin{rana2016univariate} compared two types of ANN models: univariate, that use only historical photovoltaic power values, and multivariate, that use previous photovoltaic power values and weather data. They conclude that weather data is likely to be helpful for forecasting horizons of several hours, while short term photovoltaic energy can be predicted only from previous energy data without weather information.
However, some argue that clouds analysis from sky images is potentially useful for short term solar energy forecasting~\cite{chow2011intra, huang2014analytical, lipperheide2015embedded, russo2014genetic, urquhart2013sky}. For this reason, sky images have been used in solar energy forecasting~\cite{chow2011intra, chu2015short, urquhart2013sky, yang2014solar}. In these works, the cloud cover and movement are estimated via image processing techniques, for example, the widely used red-green-blue (RGB) ratio. \citebegin{peng20153d} use a Support Vector Machine to detect clouds from multiple sky imagers. Their method requires sky images captured at multiple sites, and it achieves 26\% improvement compared with the persistence model. Cloud classification methods for whole sky images have been also used in forecasting by using either handcrafted features or machine learning methods~\cite{heinle2010automatic, taravat2015neural}. \citebegin{chu2013hybrid, marquez2013intra} use sky images for intra-hour Direct Normal Irradiance forecast instead of photovoltaic power foreast. Overall, sky images are widely used to improve the forecast performance, however, modeling cloud is one of their challenging problem.
In our work, we do not rely on explicit image processing techniques, rather we automatically learn how cloud motion and sky appearance translate into future photovoltaic power production.

\paragraph{Deep learning}

Convolutional neural networks (CNNs) are a class of deep learning techniques that have achieved success in many computer vision problems, ranging from image classification~\cite{krizhevsky2012imagenet} to image recognition~\cite{long2015fully}, motion prediction~\cite{walker2015dense} and outdoor illumination estimation~\cite{hold2017deep}. 

Recent work has shown that they can also be used for forecasting tasks from historical data. \citebegin{villegas2017learning} build a model with a combination of LSTM and encoder-decoder CNNs, which generate future frames from historical image sequences. \citebegin{xingjian2015convolutional} predict future rainfall intensity in a local region over a relatively short period of time using convolutional LSTM networks. 

\revise{Deep learning techniques have also been applied to the task of forecasting solar irradiance and/or power. For example, solar irradiance can be predicted on a per-hour, one-day ahead time horizon using deep learning~\cite{qing2018hourly,ogliari2018computational}. Future photovoltaic output can be predicted from historical photovoltaic power~\cite{abdel2017accurate}, or from weather data~\cite{gensler2016deep}, using a time horizon of 1 hour. In contrast, we tackle the challenging case of 1-minute photovoltaic power prediction, and employ ground-level images, which can be acquired using much simpler hardware than weather data.}

\section{Data}
\label{section.data}

Similar to \cite{urquhart2015development}, our methods and experiments rely on a large dataset of hemispherical HDR sky images and corresponding photovoltaic power\footnote{The dataset will shortly be available here: \url{http://rodrigo.verschae.org/skyPvCapture}.}. For completeness, we briefly summarize the data capture procedure relevant to our work below. 

\subsection{Data capture}
\label{section.data.capture}

A photovoltaic panel of $10\times6$ $\mathrm{m}^2$ area was installed on the roof of a tall building in Kyoto University, \revise{Japan}. The photovoltaic cells were installed parallel to the ground, yielding a station with a generation capacity of 2500W. 

\begin{figure}
\centering
\setlength{\tabcolsep}{0pt}
\begin{tabular}{cccc}
\includegraphics[width=.25\textwidth,trim={0 0 135.5cm 0},clip]{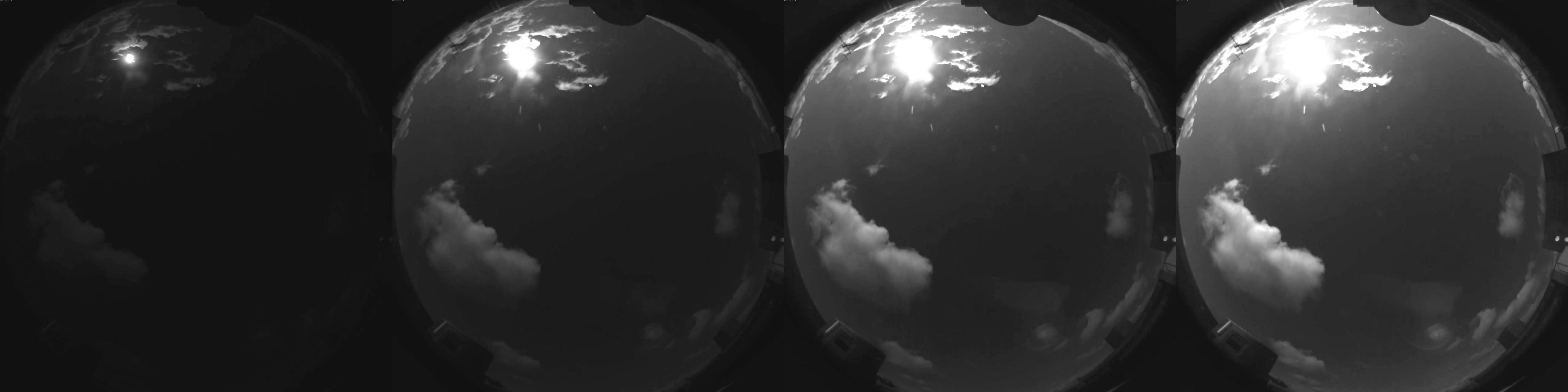} &
\includegraphics[width=.25\textwidth,trim={45cm 0 90.5cm 0},clip]{figures/pv-sample-calib-gray} &
\includegraphics[width=.25\textwidth,trim={90.5cm 0 45cm 0},clip]{figures/pv-sample-calib-gray} &
\includegraphics[width=.25\textwidth,trim={135.5cm 0 0 0},clip]{figures/pv-sample-calib-gray} \\
\revise{11ms} & \revise{88ms} & \revise{176ms} & \revise{264ms} \\
\end{tabular}
\caption{Hemispherical sky images captured at \revise{4} different exposures, \revise{shown below each image}. We only use gray-scale images to analyze the sky appearance.}
\label{data.exposure}
\end{figure}

Sky images were captured \revise{at $1280{\times}1280$ pixel resolution} using a SonyIMX265 camera equipped with a Spacecom TV1634M fisheye lens ($\mathrm{1.6mm}$ focal length) pointing to zenith. The sky camera was mounted at a distance of approximately 180~m to the center of the solar panel. A dome housing (model ASC A-SWD5VXT) was used to protect the camera.
Each second, the built-in bracketing mode in the camera was used to capture 4 images at different exposures \revise{($\{1, 8, 16, 24\} {\times} 11 = \{11, 88, 176, 264\}$ms)}. \revise{These 4 exposures allow us to capture greater dynamic range than what is possible with a single image~\cite{debevec1997recovering}, where the image intensity is proportional to the exposure time. For example, in fig.~\ref{data.exposure}, the cloud regions that appear saturated in the 264ms image are well-exposed at 88ms. However, note that the sun is always over-exposed, even in the fastest 11ms exposure. Properly exposing the sun would most likely result in improved performance (since the sun intensity is strongly correlated with the produced photovoltaic power output), but doing so requires imaging capabilities that cannot be achieved with conventional video cameras~\cite{stumpfel-afrigraph-04}}. 

The photovoltaic values from the solar panel were also recorded, where the clocks of photovoltaic capture and image capture are synchronized to an NTP server. 
This process was repeated over 1.5 years to gather a very large dataset, from which 90 days were used for the experiments reported in this paper. \revise{These 90 days were randomly sampled from the 1.5 year time period and cover different seasons. Note that while training on more data is almost certainly going to yield better results, the anticipated performance increase would probably be quite minor at the expense of significant additional training time. We thus consider these days as a representative sample of the entire dataset.}

\subsection{Data preprocessing}
\label{section.preprocessing}

Even though data was captured every second, in the remainder of the paper we will only consider data gathered every minute. Since sampling a single photovoltaic value every minute can be quite noisy, we compute the mean of all values gathered \revise{in the $[t - 60\mathrm{s}, t]$ interval, where $t$ is sampled every minute.} We remove all the data where the image is invalid (too dark), or photovoltaic power output is 0 (below sensitivity threshold). 

\revise{When image data is used, each input image actually corresponds to a 3D matrix with $4{\times}5$ channels, where we stack the images (4 exposures in gray scale) taken 60 seconds before the current time $t$ with a 15 seconds interval (therefore, 5 images are used for every minute).}

\section{Overview and notation}
\label{section.overview}

In this paper, we refer to the solar panel power output at current time $t_0$ as $p_{t_0}$. The future power value at time $t_0+t$ is thus $p_{t_0+t}$. Since we are interested in short-term forecasting (nowcasting), here $t$ is typically on the order of 1 minute (experiments on varying $t$ will be provided in sec.~\ref{sec:time-horizons}). The power variation $\Delta p_{t_0}$ between time $t_0+t$ and $t_0$ is thus given by
\newcommand{\hatdeltap}{\ensuremath{\Delta\hat{p}}}
\newcommand{\hatp}{\ensuremath{\hat{p}}}
\begin{equation}
\Delta p_{t_0} = p_{t_0+t} - p_{t_0} \,.
\end{equation}
Our goal is to learn to estimate the power variation $\hatdeltap_{t_0}$, such that the future power value $\hatp_{t_0+t}$ can be obtained with
\begin{equation}
\hat{p}_{t_0+t} = p_{t_0} + \hatdeltap_{t_0}\,.
\end{equation}
Here, the ``hat'' notation is used to identify estimated values, as opposed to ground truth values obtained from the actual panel. 

In this work, we introduce a variety of deep neural network models to estimate the variation $\hatdeltap_{t_0}$ from current and previous power values $\mathbf{p} = [ p_{t_0-k}, \ldots, p_{t_0} ]$:
\begin{equation}
\hatdeltap_{t_0} = f(\mathbf{p} ; \mathcal{W}) \,,
\label{eq.net.mlp}
\end{equation}
where the model $f(\cdot)$ is parameterized by a set of trainable weights $\mathcal{W} = [\mathbf{W}_1, \ldots, \mathbf{W}_L]$, with $L$ being the number of layers in the neural network. The neural network learns to map current and previous power outputs $\mathbf{p}_{t_0}$ to the variation $\hatdeltap_{t_0}$ with a set of non-linear functions and weights. 

We refer to the set $\mathcal{I}$ of current and previous sky images as $\mathcal{I} = [\mathbf{I}_{t_0-k}, \ldots, \mathbf{I}_{t_0}]$. A neural network which learns to map current and previous images and power outputs to the power variation $\hatdeltap_{t_0}$ is defined as 
\begin{equation}
\hatdeltap_{t_0} = f(\mathbf{p}, \mathcal{I} ; \mathcal{W}) \,.
\label{eq.net.cnn}
\end{equation}
In the following section, we present different deep learning approaches to model eq.~(\ref{eq.net.mlp}) and (\ref{eq.net.cnn}).

\section{Deep learning architectures to predict \revise{photovoltaic} production}
\label{section.methods}

\subsection{MLP with past photovoltaic values}
\label{section.mlp}

\begin{figure}
\centering
\includegraphics[width=.5\linewidth,trim={0 0 19mm 0},clip]{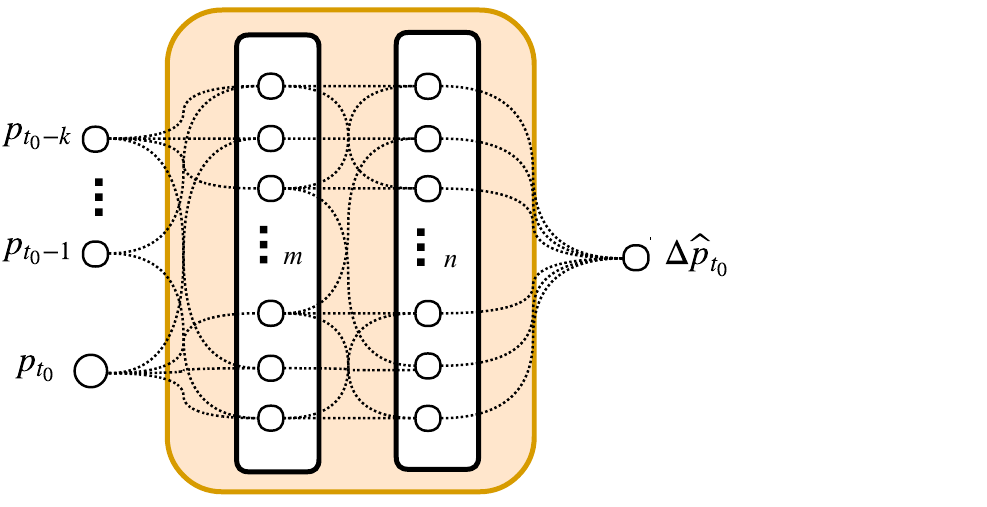}
\caption{A multilayer perceptron (MLP) $m{\times}n{\times}1$ network to learn the power variation $\hatdeltap_{t_0}$ from input historical values $p_{t_0-k}, ..., p_{t_0}$. This MLP $m{\times}n{\times}1$ network contains two hidden layers of $m$ and $n$ neurons respectively, and one neuron in the output layer. Every neuron in one layer is connected to every neuron in the preceding layer.}
\label{net.mlp}
\end{figure}

First, we define a baseline model (similar to the univariate model in~\cite{rana2016univariate}) with a multilayer perceptron (MLP) network. This network takes in only historical photovoltaic power values $\mathbf{p}$ as input to forecast the variation $\hatdeltap_{t_0}$ as in eq.~(\ref{eq.net.mlp}), and its structure is illustrated in fig.~\ref{net.mlp}. This MLP $m{\times}n{\times}1$ network is composed of two hidden layers with $m, n$ neurons in each hidden layer, and one output layer containing a single neuron. \revise{Using a separate validation dataset, it was determined that the values $m=n=64$ provided a good compromise between performance and generalization.} Batch normalization~\cite{IoffeBNA2015}, and the ``tanh'' activation function are used after all hidden layers, the ``sigmoid'' activations function is applied after output layer. This network is hereafter called the ``MLP'' model. 

This MLP model is trained to minimize the L2 difference between the estimated variation and the ground truth:
\begin{equation}
\mathcal{L}_{\Delta p_{t_0}} = ||\hatdeltap_{t_0}-\Delta{p}_{t_0}||_2 \,.
\label{eq.loss.deltap}
\end{equation}
In the following sections, we will keep using the notation MLP $m{\times}n{\times}x$ to build other neural networks with varying number of layers and neurons.

\subsection{CNN integrating sky images}
\label{section.cnn}

\begin{figure}[!ht]
\centering
\includegraphics[width=.7\linewidth,trim={12mm 8mm 18mm 0},clip]{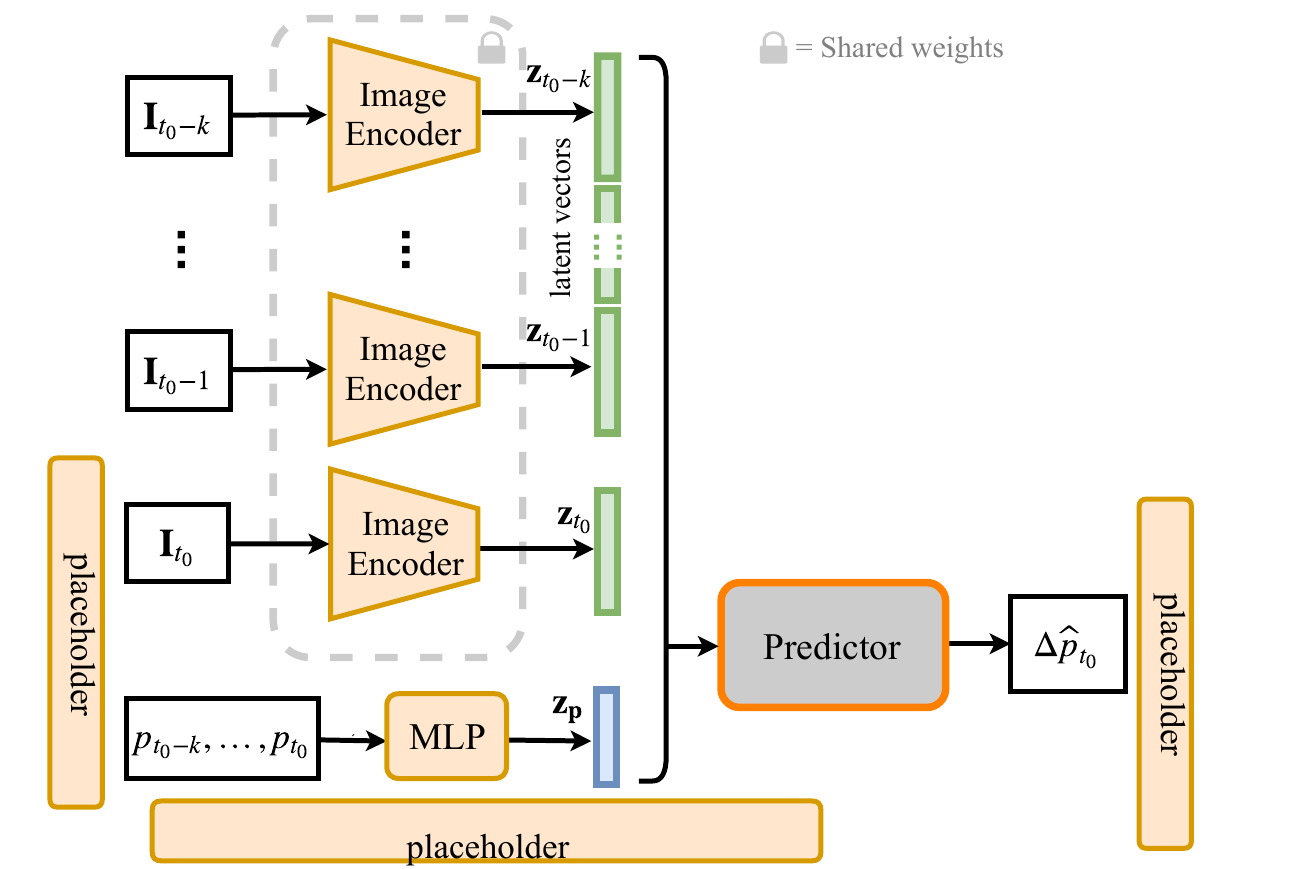}
\caption{Our deep CNN architecture, composed of an image encoder, the MLP from sec.~\ref{section.mlp}, and a predictor. The image encoder takes an image $\mathbf{I}_i$ as input and encodes it into a latent vector $\mathbf{z}_i$ (green), $i \in{\{t_0{-}k, \ldots, t_0\}}$. It uses the the same set of weights for each image. 
The historical power values $p_{t_0-k},\ldots, p_{t_0}$ are encoded into the network as another latent vector  $\mathbf{z}_\mathbf{p}$ (blue) by a MLP $m{\times}m{\times}64$ network. The latent vectors $\{\mathbf{z}_{t_0-k}, \ldots, \mathbf{z}_{t_0}\}$ and $\mathbf{z}_\mathbf{p}$ are concatenated and fed to a predictor, which is another MLP $n{\times}n{\times}n{\times}1$ network composed of 3 hidden layers. We use $m{=}64$ and $n{=}1024$ for the two MLP networks respectively.}
\label{net.cnn}
\end{figure}

While the MLP model presented above can naturally learn from previous power values $\mathbf{p}$, it cannot straightforwardly use previous images $\mathcal{I}$ associated to each $p \in \mathbf{p}$. Therefore, we propose to instead use a combination of convolutional neural networks (CNNs) with MLPs to efficiently learn from these two sources. The proposed hybrid structure, heretofore named the ``CNN'' model for simplicity, is illustrated in fig.~\ref{net.cnn}.

As shown in fig.~\ref{net.cnn}, power values $\mathbf{p}$ are integrated to the network by a MLP $m{\times}n{\times}64$. This encodes the power values into a $\mathrm{64D}$ intermediate vector $\mathbf{z}_\mathbf{p}$. 
In addition, every image $\mathbf{I}_i \in \mathcal{I}$ gets compressed into a latent vector $\mathbf{z}_i$ independently by an image encoder. The latent vectors $\mathbf{z}_\mathcal{I}=\{\mathbf{z}_{t_0-k}, \ldots, \mathbf{z}_{t_0}\}$ and $\mathbf{z}_\mathbf{p}$ are then concatenated together and fed to a predictor module. This is simply another 3-layer MLP $n{\times}n{\times}n{\times}1$ network which is expected to learn the spatial and temporal changes and estimate the power variation $\hatdeltap_{t_0}$ from these latent vectors. \revise{In this case, $m=64$ and $n=1024$. As before, a separate validation dataset was used to determined these values. }

\begin{figure}
\centering
\footnotesize
\begin{tabular}{cc}
\includegraphics[height=.17\textwidth,trim={0 0 23mm 13mm},clip]{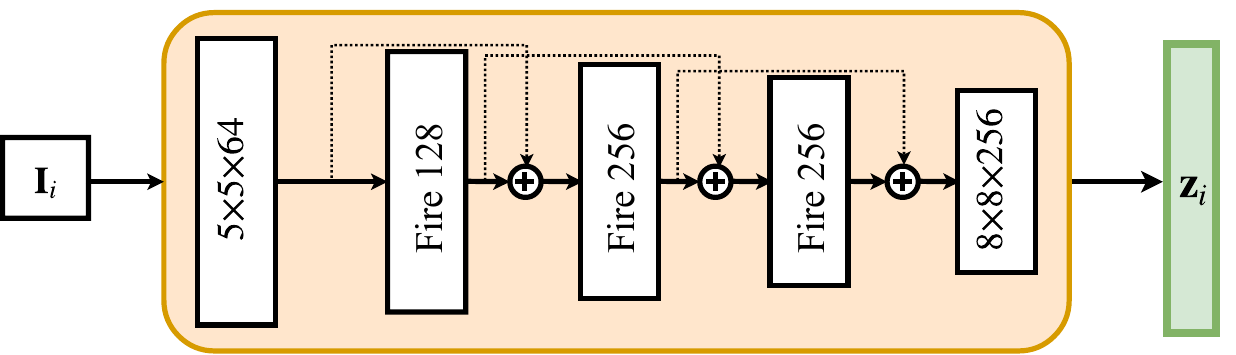} &
\includegraphics[height=.17\textwidth,trim={0 0 0 13mm},clip]{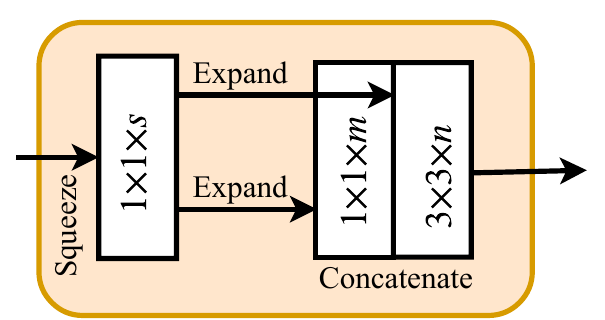} \\
(a) Image encoder & (b) Fire module
\end{tabular}
\caption{Image encoder structure. The image encoder (a) takes a 2D image  $\mathbf{I}_i$ as input and compresses it into a vector $\mathbf{z}_{i}, (i \in \{t_0{-}k, \ldots, t_0\})$. The image information is processed and passed through different layers from left to right in the encoder (arrows). The image encoder is build with a convolution layer with $5{\times}5{\times}64$ filters, then followed by three fire modules, and a $8{\times}8{\times} 256$ convolution layer. Batch normalization, ReLU activation and max pooling are used after each layer. Residual links are used across the fire modules (dash lines). The last convolution layer maps the 2D activation map into a 1D vector. 
A fire module is constructed with a squeeze convolution layer (which has only $1{\times}1$ filters), feeding into an expand layer that has a mix of $1{\times}1$ and $3{\times}3$ convolution filters, as shown in (b). We use $s{=}16, m{=}n{=}c/2$ where $c$ is the number of output channels.}
\label{net.encoder}
\end{figure}

The image encoder introduced above is a 5-layer convolutional neural network, as shown in fig.~\ref{net.encoder}. The image is first processed by a regular convolution layer, followed by 3 fire modules~\cite{iandola2016squeezenet} and a final regular convolution layer. The structure of a fire module~\cite{iandola2016squeezenet} is shown in fig.~\ref{net.encoder}(b). Batch normalization~\cite{IoffeBNA2015} and the ReLU activation function~\cite{nair2010rectified} are used in each layer. In addition, max pooling is used after each layer to reduce the spatial resolution, which allows different layers to view the input image at a different scale. Residual links~\cite{he2016deep} are used between different layers, which makes the network easier to optimize and gain accuracy. This network is trained by minimizing the loss function in eq.~(\ref{eq.loss.deltap}), just like the MLP in sec.~\ref{section.mlp}.

\subsection{LSTM for modeling temporal information}
\label{section.lstm}

The CNN from the previous section first treats every image independently and merges their intermediate representations after each has been processed. In this section, we present a structure which takes into account the temporal information present in the data. In particular, we use ``Long Short-Term Memory'' (LSTM) networks~\cite{hochreiter1997long}, a variation of recurrent networks (RNN). LSTMs are widely used in processing temporal data, such as precipitation forecasting~\cite{xingjian2015convolutional}, future frame prediction~\cite{srivastava2015unsupervised}, and video to text~\cite{venugopalan2015sequence}, etc. \revise{More closely related to our work, LSTMs have also been used for photovoltaic forecasting from weather data~\cite{gensler2016deep}, from historical photovoltaic power~\cite{abdel2017accurate} or for day-ahead solar irradiance prediction~\cite{qing2018hourly,ogliari2018computational}.} Unlike stateless CNNs, LSTM networks contain loops which allow sequential states to be memorized. 

\begin{figure}[ht]
\centering
\footnotesize
\begin{tabular}{c}
\includegraphics[width=.99\textwidth,trim={11mm 8mm 18mm 0},clip]{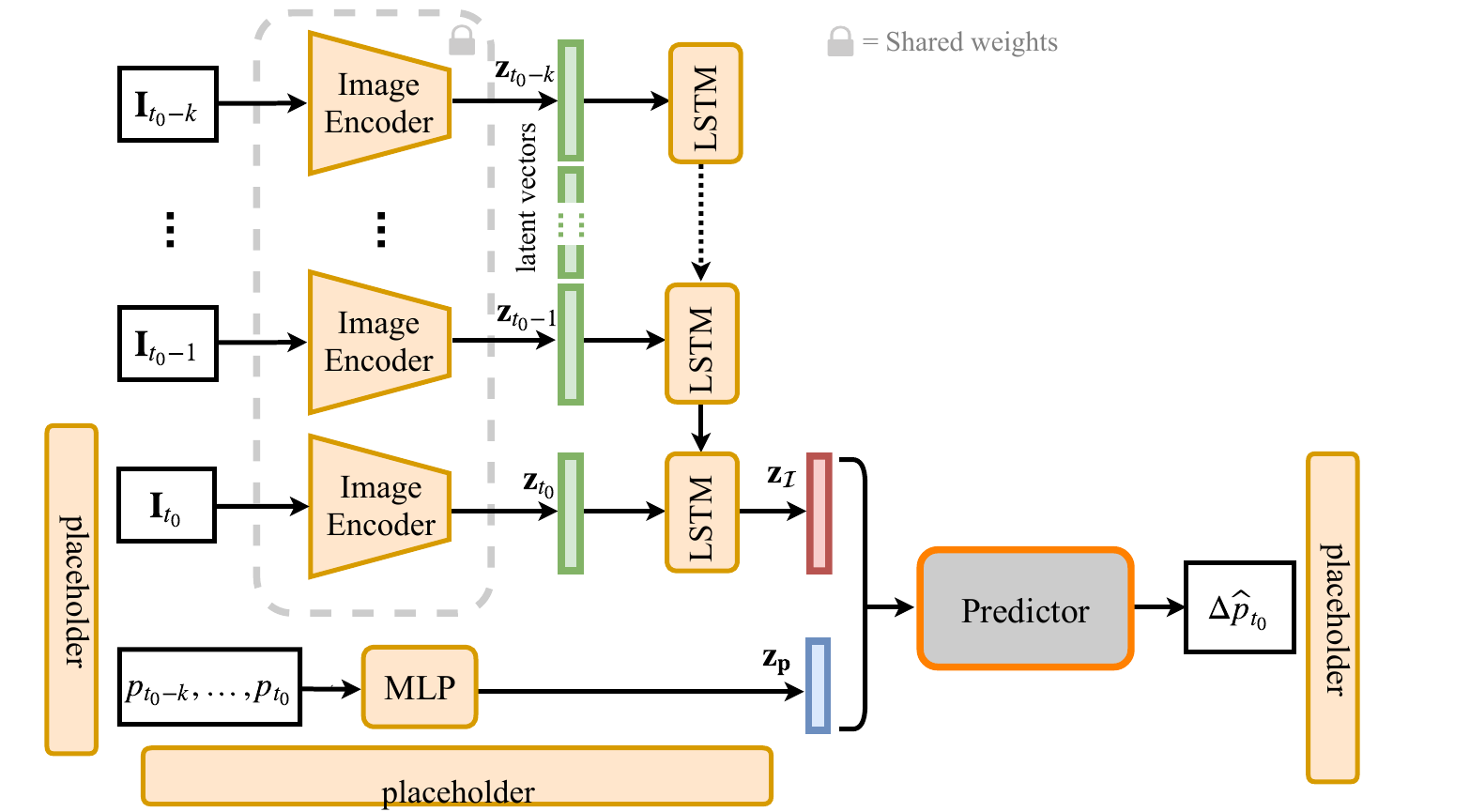}
\end{tabular}
\caption{LSTM structure. The latent vectors $\{\mathbf{z}_{i}|i = {t_0{-}k, \ldots, t_0}\}$ (green), are generated from the image encoder, and fed iteratively to a LSTM block. The (repeating) LSTM module is a 2-layer network, which maps a pair of state and input to a pair of state and output (the vertical arrows indicate the state flow from one step to another).}
\label{net.lstm}
\end{figure}

In this network, instead of concatenating the latent vectors $\{\mathbf{z}_{i}|i = {t_0{-}k, \ldots, t_0}\}$ that are encoded from the input image set $\mathcal{I}$ as with the CNN in sec.~\ref{section.cnn}, we instead use a 2-layer LSTM network on these latent vectors as illustrated in fig.~\ref{net.lstm}. The recurrent connection in the LSTM layer is useful to capture the structure of sequences~\cite{hermans2013training}. The output vector $\mathbf{z}_\mathcal{I}$ in the last step should therefore encode temporal information from all the past latent vectors. It is then concatenated with $\mathbf{z}_\mathbf{p}$ and fed to a predictor sub-network, which outputs the predicted variation $\hatdeltap_{t_0}$. The same loss function as in sec.~\ref{section.mlp} and \ref{section.cnn}, eq.~(\ref{eq.loss.deltap}), is used to train this network. \revise{For simplicity, we refer to this network with the ``LSTM'' shorthand from now on (even if, strictly speaking, it is not solely an LSTM module).}

\subsection{Multi-task learning}
\label{section.branches}

\newcommand{\fullmodel}{LSTM-Full}
\begin{figure}[ht]
\centering
\begin{tabular}{cc}
\includegraphics[height=.25\textwidth,trim={5mm 8mm 110mm 0mm},clip] {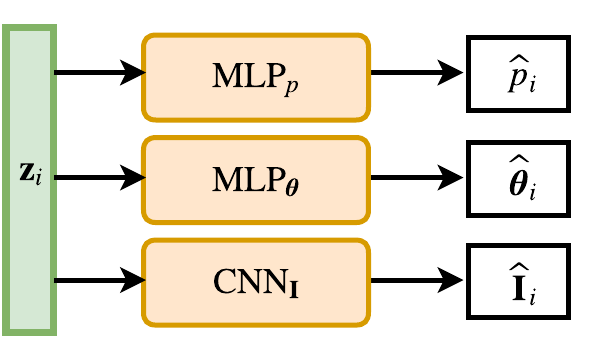} &
\includegraphics[height=.25\textwidth,trim={75mm 8mm 38mm 0mm},clip] {figures/structure-regressor-predictor}\\
(a) Regressors & (b) Predictors
\end{tabular}
\caption{Additions to the network for multi-task learning. The regressors take the latent vector $\{\mathbf{z}_{i}|i = {t_0{-}k, \ldots, t_0}\}$ as input; the predictors take the vector from LSTM $\mathbf{z}_\mathcal{I}$ as input. The regressors contain three different tasks: sun position regression, photovoltaic energy regression and image reconstruction for every input latent vector. The sun position regression and photovoltaic energy regression are MLP $256{\times}2$ (azimuth and elevation) and MLP $256{\times}1$ network respectively. The image reconstruction branch is a CNN network, stacked with up-sampling and convolution layers. The predictors are three independent MLP networks to predict the power variation $\hatdeltap_{t_0}$, sun position variation (MLP $256{\times}2$), and sky intensity variation (MLP $256{\times}1$).
Note that the regressors are used to estimate the absolute values from every latent vector in $\{\mathbf{z}_i|i=t_0{-}k, \ldots, t_0\}$, the predictors are used to forecast the variations for the future based on the single LSTM output vector $\mathbf{z}_\mathcal{I}$. }
\label{net.branches}
\end{figure}

Since predicting the power output variation $\hatdeltap_{t_0}$ is our main task, so far, all the networks have been trained to minimize the error in predicting $\hatdeltap_{t_0}$. However, it has been shown that \revise{better performance can be achieved} when simultaneously training for multiple, related, tasks.~\citebegin{zhang2014facial} use multi-tasking learning to improve robustness for facial landmark detection by adding auxiliary tasks, such as age estimation, gender classification, etc.~\citebegin{girshick2016region} train an end-to-end object classification network with a multi-task loss to improve the classification accuracy. This is commonly known as multi-task learning~\cite{caruana1998multitask}. In this section, we introduce several related tasks and associated loss functions, which can help the network to achieve better performance. While we could technically ask any of the previously-introduced network to perform these new tasks, we focus on the LSTM network here. The resulting combined network will be dubbed ``LSTM-Full'' in the remainder of the paper. 

\begin{table}
\centering
\begin{tabular}{llcc}
\toprule
Name & Type & Symbol & Loss \\ 
\toprule
Absolute power output & instant & $p_i$ & $\mathcal{L}_{p} = \sum_{i=t_0{-}k, \ldots, t_0} ||\hat{\mathbf{p}}_i-\mathbf{p}_i||_2$  \\
\midrule
Sun position & spatial & $\boldsymbol{\theta}_i$ & $\mathcal{L}_\theta = \sum_{i=t_0{-}k, \ldots, t_0} ||\hat{\boldsymbol{\theta}}_i-\boldsymbol{\theta}_i||_2$ \\
Image & spatial & $\mathcal{I}_i$ & $\mathcal{L}_\mathcal{I} = \sum_{i=t_0{-}k, \ldots, t_0} ||\hat{\mathcal{I}}_i-\mathcal{I}_i||_2$ \\
\midrule
Sun position variation & temporal & $\Delta\boldsymbol{\theta}_i$ & $\mathcal{L}_{\Delta\theta} = ||\Delta\hat{\boldsymbol{\theta}}-\Delta\boldsymbol{\theta}||_2$ \\
Sky intensity variation & temporal & $\Delta\mathbf{s}_i$ & $\mathcal{L}_{\Delta s} = ||\Delta\hat{\mathbf{s}}-\Delta\mathbf{s}||_2$ \\
\bottomrule
\end{tabular}
\caption{Summary of the additional losses for multi-task learning. The ``instant'' and ``spatial'' tasks are computed from all $K$ latent vector in $\{\mathbf{z}_i|i=t_0{-}k, \ldots, t_0\}$, while the ``temporal'' tasks are computed from the single LSTM output vector $\mathbf{z}_\mathcal{I}$.} 
\label{tab.multi-task}
\end{table}

Table~\ref{tab.multi-task} presents an overview of the different sub-tasks and their associated losses that are used to train the network, and fig.~\ref{net.branches} illustrates the corresponding deep learning architecture used to predict each of these sub-tasks. Overall, there are three types of tasks: ``instant'' and ``spatial'', which can get predicted from the latent vectors $\mathbf{z}_i$; and ``temporal'' which get predicted from $\mathbf{z}_\mathcal{I}$. 

The sole ``instant'' sub-task is that of regressing the \emph{absolute} power output $p_i$ at each timestep $i$. Note that while regressing this value is typically harder than predicting the variation (sec.~\ref{section.overview}), we hypothesize that regressing this value from the image encoder alone should make it model the relationship between sky appearance and power output more accurately. 

We add two ``spatial'' sub-tasks. First, we train a $\text{MLP}_{\boldsymbol{\theta}}$ to predict the sun position in the image ($\boldsymbol{\theta}$ representing the 2-vector of elevation and azimuth in spherical coordinates). In addition, we also ask the network to predict the entire sky image $\mathcal{I}_i$ itself. This is done through a CNN, which is composed of 5 convolution layers. Batch normalization, ReLU activation and upsampling are used after each layer.

Finally, two ``temporal'' sub-tasks are also added. First, the sun position variation $\Delta \boldsymbol{\theta}_i$ is estimated, which is simply the difference between two sun positions (computed on the elevation and azimuth independently). Second, the sky intensity variation is also estimated. Here, the sky intensity $\mathbf{s}_i$ is the integral of the high dynamic range sky image
\begin{equation}
s = \frac{1}{2\pi}\sum_{i}^{N}{b_i\Delta\boldsymbol{\omega}_i}  \,, 
\end{equation}
where $N$ is the number of pixels, $b_i$ is the pixel intensity, $\omega_i$ is the solid angle spanned by pixel $i$. The HDR image is computed \revise{by combining the 4 exposures using the method of \citebegin{debevec1997recovering}}.

We use the following loss function to train the \fullmodel~network:
\begin{equation}
\begin{aligned}
\mathcal{L} = \mathcal{L}_{\Delta p} + \lambda_{\Delta\theta}\mathcal{L}_{\Delta\theta} + \lambda_{\Delta s}\mathcal{L}_{\Delta s} + \lambda_{p}\mathcal{L}_{p} + \lambda_{\theta}\mathcal{L}_\theta + \lambda_{I}\mathcal{L}_I \,,
\end{aligned}
\end{equation}
where the $\lambda_*$ are scale factors that balance the gradient that flows to the image encoder. In our experiments, we use $\lambda_{\Delta\theta}=10^3$, $\lambda_{\Delta s}=10^{-3}$, $\lambda_{\theta}=\lambda_{p}=\lambda_{I}=0.1$.

\subsection{Network training}
\label{section.train}

The networks are trained \revise{and evaluated} on a large dataset \revise{gathered using the procedure described in sec.~\ref{section.data}}. \revise{The entire dataset contains 90 days, which is split randomly (by days): 80\% (72 days) for training the networks, and 20\% (18 days) for test}. Every network is trained in a supervised way, where known targets are available at training time.

We train the networks using the ADAM~\cite{kingma2014adam} optimizer with a learning rate of $10^{-3}$ for the image encoder and $30^{-4}$ for the other layers. At test time, we only give the past and current power values with the corresponding sky images to predict the future power value.
Training 200 epochs takes roughly 24 hours on an Nvidia Titan X GPU. At test time, inference takes approximately 70ms.

\section{Experiments}
\label{section.experiments}

In this section, we first introduce the error metrics used for model performance evaluation and comparison. Then we quantitatively compare different models for 1-min forecasting task over different metrics and different weather conditions. Afterwards, we show qualitative results on 1-min forecasting. Finally, we evaluate the performance for different exposures and different forecast horizons.

\subsection{Pre-processing}

To ease training, we pre-process the power values with the following procedure. First, the historical power values $\mathbf{p}$ and variations $\hatdeltap_{t_0}$ used in any of the deep learning models $f(\cdot)$ (see eqs.~(\ref{eq.net.mlp}) and (\ref{eq.net.cnn})) are converted to log space according 
\begin{equation}
 g(x) = 
  \begin{cases} 
      x & x < 1 \\
      \log(x) & x\geq 1 
   \end{cases}\,,
   \label{eq.data.log}
\end{equation}
and scaled by a factor of $\alpha$ to normalize the data in the $[0, 1]$ interval. The output from $f(\cdot)$ can be easily linearized by applying the inverse of eq.~(\ref{eq.data.log}).

\subsection{Error metrics}

As is commonly done in the literature~\cite{huang2014analytical,marquez2013intra,rana2016univariate,soubdhan2016robust,voyant2017machine}, we use the assumption that the future power will remain unchanged over the forecasting horizon ($\hatdeltap_{t_0} = 0$) as a baseline. This is commonly known as the ``persistence'' model. 
Our test dataset is manually split into three different categories based on the weather conditions: clear, partly cloudy and overcast. Here, we \revise{manually} define ``clear'' days as those with less than approximately 10\% clouds on average throughout the day, ``partly cloudy'' have between 10\% and 90\%, and ``overcast'' are completely overcast days. We report performance with the following metrics on each of these categories. 

The ``mean absolute error'' (MAE) measures the average absolute deviation of the estimated values from their measured values, which is less sensitive to outliers than the widely-used ``root mean squared error'' (RMSE):
\begin{equation*}
\mathrm{MAE} = \frac{1}{N}\times\sum_{i=1}^{N}{|p_i - \hat{p}_i|}\,, \quad
\mathrm{RMSE} = \sqrt{\frac{1}{N}\times\sum_{i=1}^{N}{\left(p_i - \hat{p}_i\right)^2}}\,.
\end{equation*}

The ``forecast skill score'' (SS) is used to compare performance between two methods, it is given by:
\begin{equation*}
\mathrm{SS} = \left(1 - \frac{\mathcal{E}_\text{prediction}}{\mathcal{E}_\text{baseline}} \right) \times 100\%\,,
\end{equation*}
where $\mathcal{E}_*$ is any error metric that is used to evaluate performance for every model. If the ``prediction'' model performs equally well as the ``baseline'' model, the skill score will be 0. A higher skill score thus means that the ``prediction'' model outperforms the ``baseline'' model.

We will use the skill score (SS-MAE, SS-RMSE) to compare the performance between different models. For the skill score, the baseline is always the persistence model as presented above. The absolute error metrics (MAE, RMSE) are also used to report the performance for specific models.
In our experiments, each model is trained separately and the best-performing model (in validation) is selected from 200 epochs of training.

\subsection{Comparison between different models}
\label{section.experiments.comparison}

In this experiment, we compare the different structures with respect to the persistence model. First, we compare the ``MLP'' (sec.~\ref{section.mlp}, ``CNN'' (sec.~\ref{section.cnn}) and ``LSTM'' (sec.~\ref{section.lstm}) models, to analyze whether incorporating image (CNN) and temporal (LSTM) information helps. Then, the following experiment shows that adding different subtasks could improve the photovoltaic forecasting task.

\begin{figure}[!t]
\centering
\footnotesize
\begin{tabular}{cc}
\includegraphics[width=.48\textwidth,trim={5mm 3mm 0 10mm},clip]{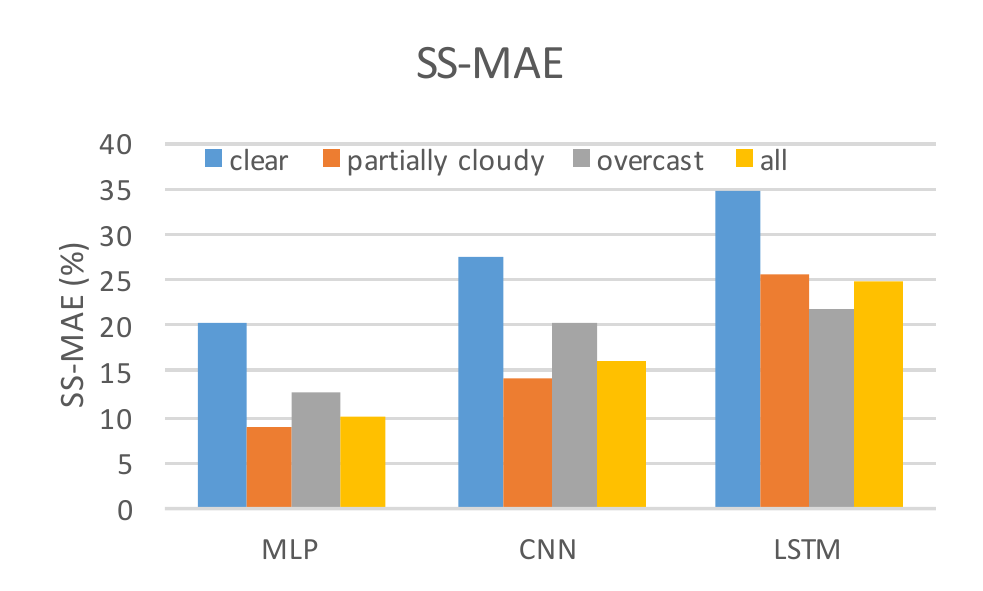} &
\includegraphics[width=.48\textwidth,trim={5mm 3mm 0 10mm},clip]{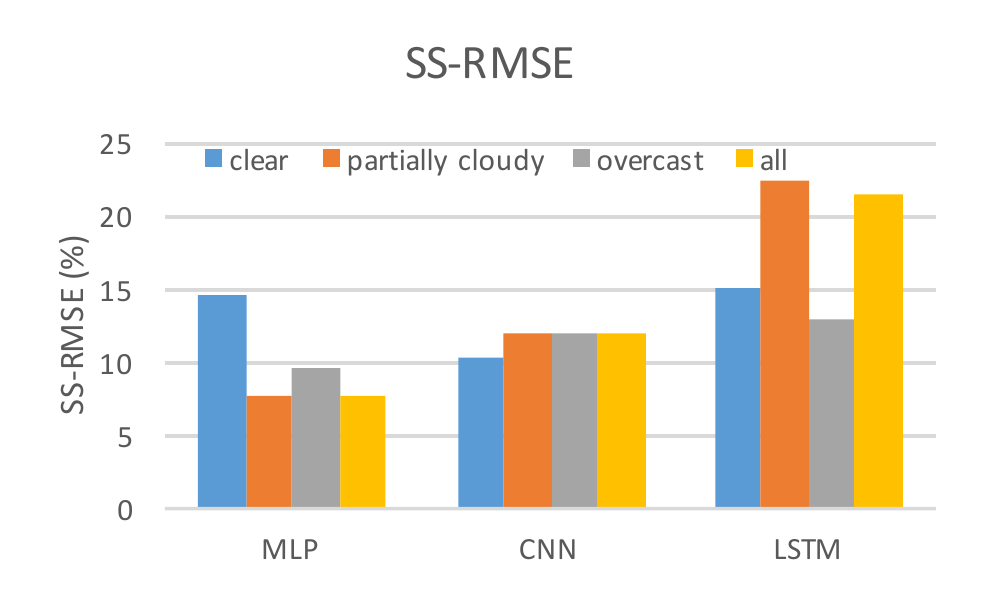} \\
\end{tabular}
\caption{Skill score compared to baseline. For every model, we compute the skill score (the higher the better) with respect to the persistence model on different weather conditions. The proposed models perform better than the persistence model over SS-MAE (left) and SS-RMSE (right). Incorporating images into the forecast model with CNNs greatly helps the performance, especially on the partially cloudy weather condition. The LSTM captures both spatial and temporal information from the sky images.}
\label{result.ss.baseline}
\end{figure}

\paragraph{Comparison between the ``MLP'', ``CNN'' and ``LSTM'' models}

We first compare the three different structures proposed in sec.~\ref{section.methods} to the persistence model on the 1-min forecasting task. The skill scores obtained are shown in fig.~\ref{result.ss.baseline}. First, we observe that all models significantly outperform the baseline persistence model. More interestingly however, including image data (``CNN'') does indeed help in predicting the future power produced by a solar panel, since the skill score is greater than that of the ``MLP'' model for all three weather conditions. However, the greatest gain is obtained by explicitly modeling the temporal nature of the problem with the ``LSTM'' model. Indeed, the ``LSTM'' model achieves SS-MAE of $34.5$, $25.5$, $21.8$ and SS-RMSE of $15.0$, $22.3$, $12.8$ on clear, cloudy, and overcast weather conditions respectively.

\paragraph{Multi-task learning}

\begin{figure}[!t]
\centering
\footnotesize
\begin{tabular}{cc}
\includegraphics[width=.48\textwidth,trim={5mm 3mm 0 10mm},clip]{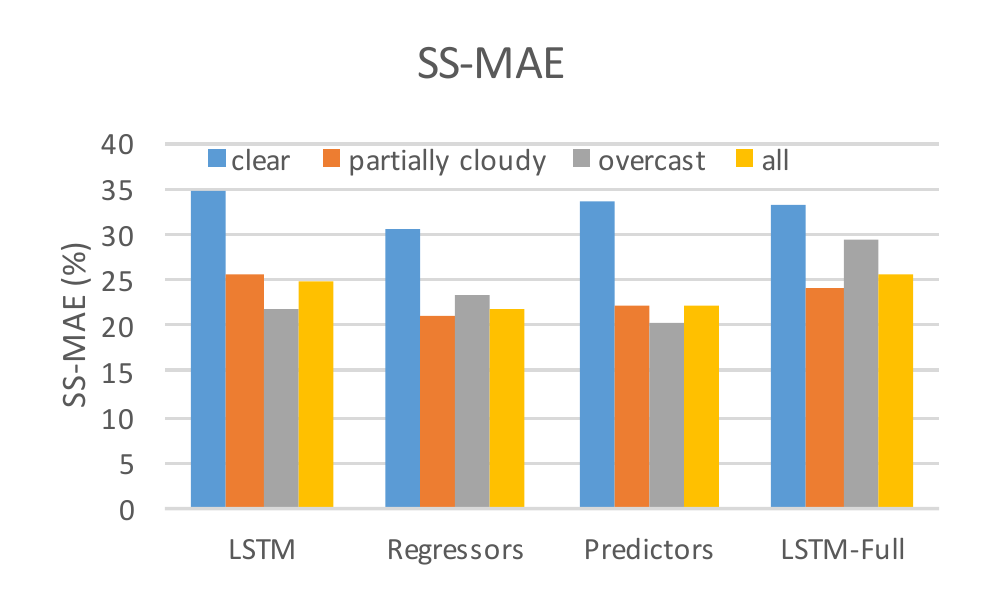} &
\includegraphics[width=.48\textwidth,trim={5mm 3mm 0 10mm},clip]{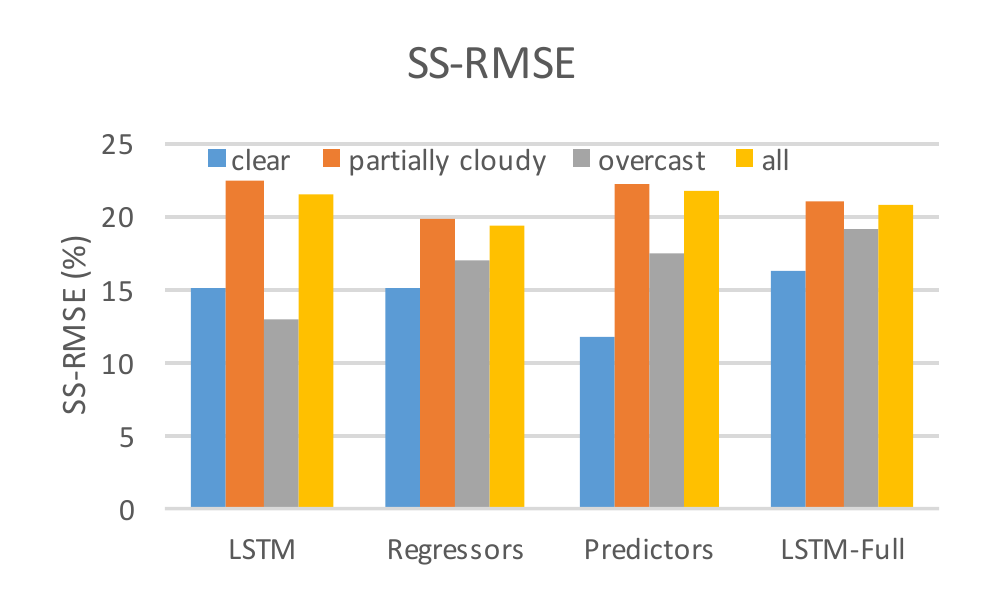} \\
\end{tabular}
\caption{The impact of learning sub-tasks to power output prediction. 
First, we add the regressors and predictors independently, then we add all the branches to the ``LSTM'' structure to build our ``\fullmodel'' network. Adding all branches helps improving the performance on clear and overcast conditions.}
\label{result.ss.branch}
\end{figure}

As proposed in sec.~\ref{section.branches}, we add regressors (for the ``instant'' and ``spatial'' sub-tasks, see table~\ref{tab.multi-task}) and predictors (for the ``temporal'' sub-tasks) to the ``LSTM'' structure respectively, to obtain the ``\fullmodel'' model. Fig.~\ref{result.ss.branch} shows that the performance is similar to the ``LSTM'' model on the SS-MAE metric. \revise{The main difference we observe is that the ``LSTM-Full'' seems to obtain a performance that is better-balanced across the different types of weather conditions. This can be valuable if the model is to be used on another dataset which might contain a different distribution of weather conditions.}

\paragraph{Results summary} Table~\ref{result.table.1min} summarizes the MAE and RMSE for the 1-minute horizon prediction for all models. For example, the ``\fullmodel'' model achieves a $5.6$, $109.3$ and $36.4$ on MAE metric and $15.3$, $203.5$ and $76.5$ on RMSE metric. For simplicity and because it is the best-performing model, subsequent experiments are conducted with ``\fullmodel'' network. 

\begin{table}[!t]
\centering
\caption{Prediction for 1-min future. All metrics are reported in watts. }
\footnotesize
\begin{tabular}{lcccccccc}
    \toprule
    \multirow{2}{*}{Model} &
      \multicolumn{2}{c}{clear} &
      \multicolumn{2}{c}{partially cloudy} &
      \multicolumn{2}{c}{overcast}  &
      \multicolumn{2}{c}{all} \\
      & {MAE} & {RMSE} & {MAE} & {RMSE} & {MAE} & {RMSE} & {MAE} & {RMSE} \\
      \midrule
    Persistence & 8.4 & 18.3 & 144.2 & 257.6 & 51.7 & 94.2 & 81.6 & 177.5 \\
    MLP         & 6.7 & 15.6 & 131.5 & 238.6 & 45.8 & 85.4 & 73.4 & 163.7 \\
    CNN         & 6.1 & 16.4 & \revise{123.5} & 227.9 & 41.2 & 83.6 & 68.6 & 156.4 \\
    LSTM        & \textbf{5.5} & 15.5 & \textbf{107.2} & \textbf{200.6} & 40.8 & 82.8 & 61.1 & \textbf{139.3} \\
    \fullmodel~ & 5.6 & \textbf{15.3} & 109.2 & 203.1 & \textbf{36.1} & \textbf{76.9} & \textbf{60.7} & 140.5 \\
    \bottomrule
  \end{tabular}
\label{result.table.1min}
\end{table}

\subsection{Qualitative visualization} 

\begin{figure}[!t]
\centering
\footnotesize
\includegraphics[width=1\textwidth,trim={2mm 6mm 5mm 0},clip]{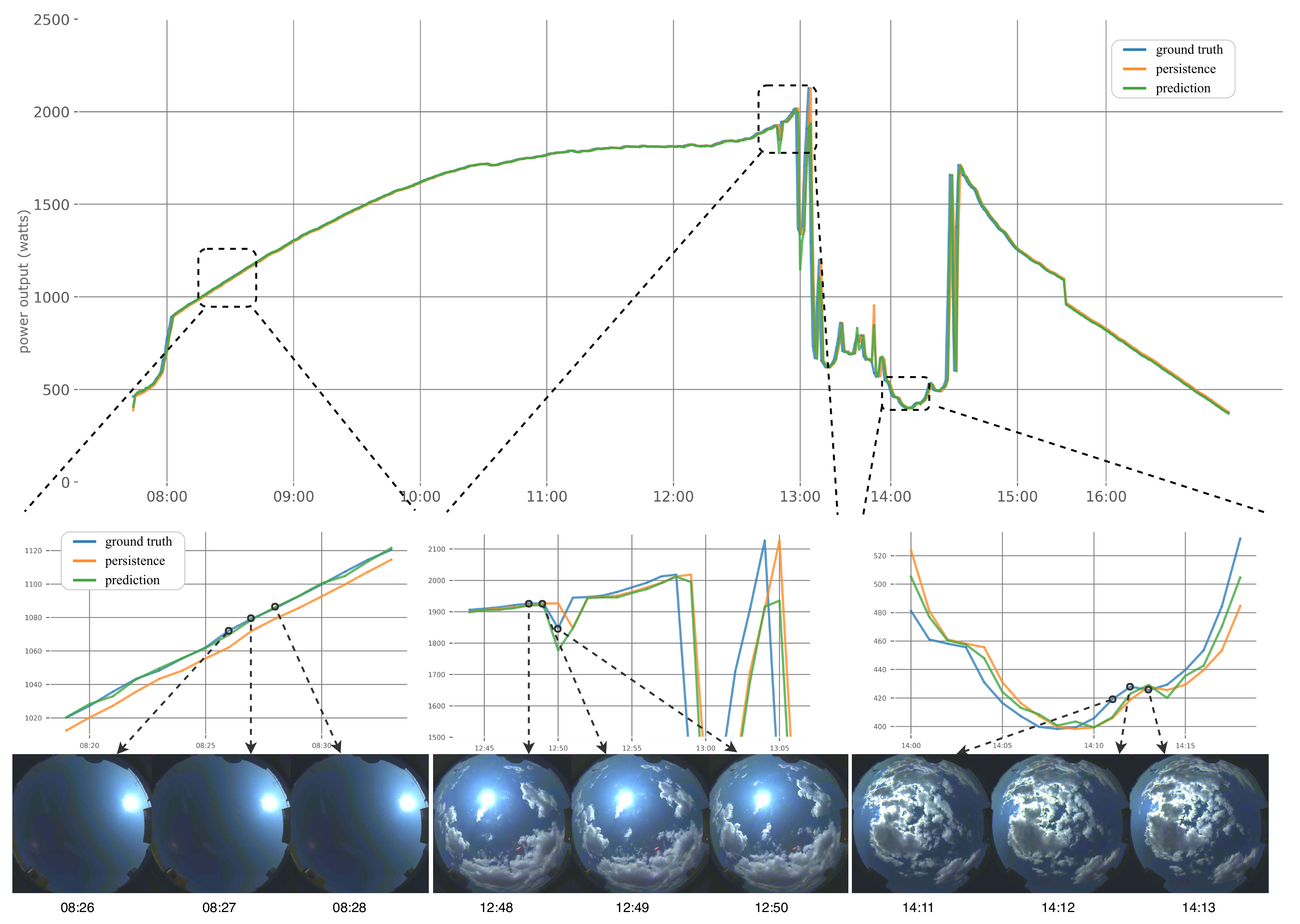}
\caption{Prediction curve for 1-min horizon on a typical day captured on 04/02/2017. 
The plot in the top shows the photovoltaic power for ground truth (blue), persistence (orange), and prediction (green). Close-up views of the curves is shown the windows in the middle. 
Three sky images are shown in the bottom for each window, their timestamps are linked to the curve (color images are only used for visualization purpose). 
(Better view in color, please zoom-in for details.)}
\label{result.curves}
\end{figure}

\begin{figure}[!t]
\centering
\footnotesize
\includegraphics[width=1\textwidth,trim={2mm 6mm 5mm 0},clip]{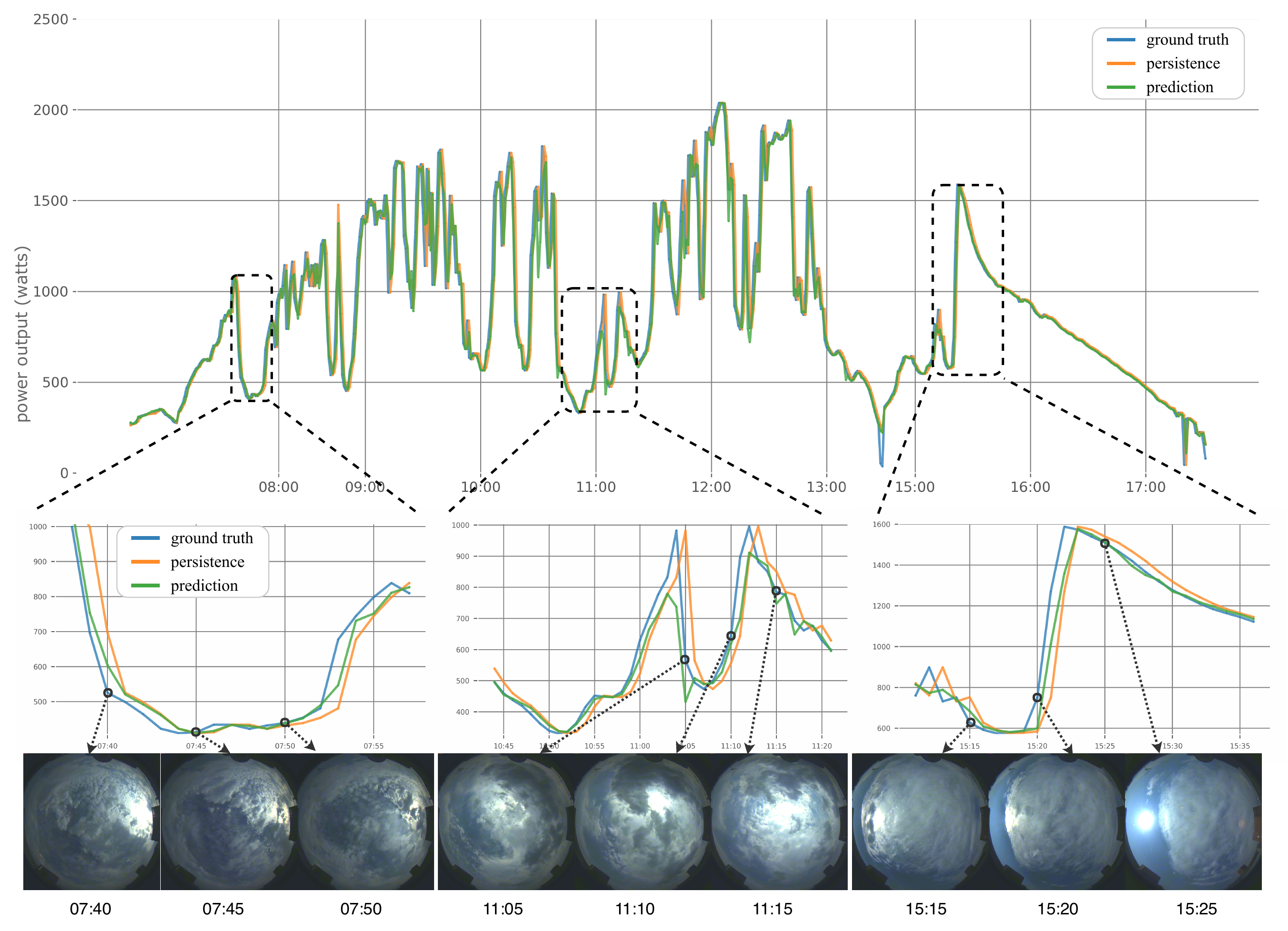}
\caption{Prediction curve for 1-min horizon on a typical day captured on 04/29/2017. 
The plot in the top shows the photovoltaic power for ground truth (blue), persistence (orange), and prediction (green). Close-up views of the curves is shown the windows in the middle. 
Three sky images are shown in the bottom for each window, their timestamps are linked to the curve (color images are only used for visualization purpose). 
(Better view in color, please zoom-in for details.)}
\label{result.curves.2}
\end{figure}

We now show qualitative results obtained with the ``\fullmodel'' model.
We plot the prediction curves for two typical days to visualize the input and prediction in fig.~\ref{result.curves} and~\ref{result.curves.2}. 

Fig.~\ref{result.curves} shows a day that is mostly clear in the morning, and cloudy in the afternoon. In the first half of the day, the model predicts the future power values very accurately (left). While sudden, unanticipated changes currently cannot be handled (middle), the model recovers when changes are less abrupt (right), even if the cloud conditions are very challenging. 

Fig.~\ref{result.curves.2} shows a more challenging case of very dynamic cloud movement causing sudden changes in output power. In this case, the model is still able to adapt to the various conditions \revise{in most cases}, even if cloud appearance vastly differs from one time instance to the next. \revise{Although} sudden rises in power output is \revise{is not predicted at the exact right time by the model (see right-most example in fig.~\ref{result.curves.2} for example), the network typically recovers very quickly.}

\subsection{Impact of different exposures}

In our experiments, the image input is the 4 exposures stacked as different channels. Using different exposures is important, since they each capture different sky components, \revise{as explained in sec.~\ref{section.data}}. For example, the first exposure (shortest, \revise{of 11ms}) in fig.~\ref{data.exposure} captures the brightest sun and its surroundings, while the last exposure (longest, \revise{of 264ms}) makes the clouds visible in the sky image. 
\revise{To validate this intuition empirally}, we \revise{perform two experiments where the image input is replaced with a \emph{single} exposure, rather than all four. To do so, we train the ``\fullmodel'' model both the shortest and fastest exposure, and compare results with all exposures in table~\ref{result.table.exposure}.} \revise{This experiment shows} that using different exposures indeed helps the photovoltaic forecasting, where MAE \revise{of} 5.6, 109.3 and 36.4 \revise{were obtained} for clear, partially cloudy and overcast weather conditions.

\begin{table}[!t]
\centering
\caption{Prediction for different exposures. All metrics are reported in watts.}
\setlength{\tabcolsep}{0.6em}
\footnotesize
\begin{tabu}{lcccccccc}
    \toprule
    \multirow{2}{*}{Exposure} &
      \multicolumn{2}{c}{clear} &
      \multicolumn{2}{c}{partially cloudy} &
      \multicolumn{2}{c}{overcast} &
      \multicolumn{2}{c}{all} \\
      & {MAE} & {RMSE} & {MAE} & {RMSE} & {MAE} & {RMSE}  & {MAE} & {RMSE} \\
      \midrule
    shortest & 6.9 & 16.9 & 117.6 & 215.2 & 44.3 & 87.9 & 68.7 & 155.6 \\
    longest & \textbf{5.3} & 16.2 & 113.4 & 210.9 & 36.8 & \textbf{74.5} & 66.8 & 151.9 \\
    \revise{all} & 5.6 & \textbf{15.3} & \textbf{109.3} & \textbf{203.5} & \textbf{36.4} & 76.5 & \textbf{60.7} & \textbf{140.5} \\
    \bottomrule
  \end{tabu}
\label{result.table.exposure}
\end{table}

\subsection{Varying the time horizons}
\label{sec:time-horizons}

We now \revise{explore the applicability of our method to longer time horizons. In particular, we experiment with 2-, 5- and 10-min future prediction and mirror the scheme used in the 1-minute case.}

\revise{First, as in sec.~\ref{section.preprocessing}, the raw data is filtered by assigning the data at each $x$ minute as being the average over the raw data on the $[t-x, t]$ interval. For a prediction $x$ minutes in the future, we consider historical images and photovoltaic power values sampled at each $x$ minute in the past, for a total time horizon of $5x$ minutes. For example, the 2-min horizon experiment takes as input data sampled every 2 minutes, over a 10-minute past time window.}

\revise{Results are reported in table~\ref{result.table.horizon}. We note that, even with longer time horizons, our approach still achieves low error under clear skies and outperforms the persistence model with an SS-MAE of $16.4\%$ and SS-RMSE of $11.5\%$ for 2-min horizon, an SS-MAE of $14.4\%$ and SS-RMSE of $10.4\%$ for 5-min horizon, and an SS-MAE of $12.1\%$ and SS-RMSE of $7.7\%$ for 10-min horizon. We note however that, as time horizon increases, the skill score decreases. One potential reason for this is that sky appearance can change dramatically over a longer horizon, especially for partially cloudy and overcast skies. For longer time horizon forecast, sky images may not actually be a reliable data source for predicting these changes in weather conditions. Combining other data inputs such as NWP and satellite images could potentially help to predict longer time horizons.}

\begin{table}[!t]
\centering
\caption{Prediction for different horizons. All metrics are reported in watts.}
\footnotesize
\setlength{\tabcolsep}{0.35em}
\begin{tabu}{lcccccccccc}
    \toprule
    \multirow{2}{*}{Horizon} &
      \multicolumn{2}{c}{clear} &
      \multicolumn{2}{c}{partially cloudy} &
      \multicolumn{2}{c}{overcast} &
      \multicolumn{4}{c}{all} \\
      \rowfont{\scriptsize}
      & {MAE} & {RMSE} & {MAE} & {RMSE} & {MAE} & {RMSE} & {MAE} & {RMSE} & {SS-MAE} & {SS-RMSE} \\
      \midrule
    1-min & 5.6 & 15.3 & 109.3 & 203.5 & 36.4 & 76.5 & 60.7 & 140.5 & 25.5\% & 20.8\% \\
    \revise{2-min} & 9.1 & 20.6 & 160.7 & 263.2 & 54.6 & 92.6 & 90.2 & 181.5 & 16.4\% & 11.5\% \\
    \revise{5-min} & 15.2 & 30.4 & 203.0 & 292.4 & 87.4 & 126.5 & 120.7 & 206.3 & 14.4\% & 10.4\% \\
    \revise{10-min} & 21.4 & 36.8 & 239.1 & 321.8 & 133.7 & 183.9 & 153.8 & 238.5 & 12.1\% & 7.7\% \\
    \bottomrule
  \end{tabu}
\label{result.table.horizon}
\end{table}

\section{\revise{Discussion of limitations}}
\label{section.discussion}

\revise{
One of the main limitations of this work is the reliance on data coming from a single site, captured by a single camera and a single photovoltaic panel. While this limitation is somewhat offset by the fact that our dataset contains a rich temporal sampling of the site, thus yielding a varied set of weather conditions, this begs the question of whether or not months of data will always be needed to re-train our approach on another site. We conjecture that, since we train on photovoltaic \emph{changes} (and not absolute values), entirely re-training the model on another site could potentially be avoided and instead, our pre-trained networks could be used as starting point and fine-tuned on another dataset (to adapt to the photovoltaic panel and camera characteristics, as well as potentially different weather conditions). This fine-tuning strategy has been demonstrated time and again in the literature (see \cite{Girshick2013} for a well-known example), and would be a potential solution to address this issue. 
}

\revise{
We also note that there is little difference in overall performance between our ``LSTM'' and ``LSTM-Full'' models. As discussed in sec.~\ref{section.experiments.comparison}, we do however note that the ``LSTM-Full'' model obtains a performance that is better-balanced across the different types of weather conditions. In addition, the auxiliary tasks in table~\ref{tab.multi-task}---such as the sun position, the sky image, the sun position variation and the sky intensity variation---are \emph{independent} from the solar panel. Evidence from the literature~\cite{devin2017learning}, which demonstrates that networks trained on multiple tasks have can better generalize on other datasets, leads us to believe that the ``LSTM-Full'' model may have a better chance of adapting to another site, for example with the same fine-tuning strategy mentioned above.
}

\revise{
Another limitation is that our models have difficulty predicting very sharp changes in photovoltaic output, such as those created by a sudden cloud moving in front of the sun. For example, in the right-most example of fig.~\ref{result.curves.2}), the model is not able to anticipate the exact time at which the photovoltaic power starts to rise (around 15:19). The sudden rise in photovoltaic production is due to the fact that the large clouds move away from the sun at that exact time. The fact that the network cannot predict exactly when this happens is probably due to the fact that there are, in fact, very little difference between the images around that time, and thus the visual cue is not sufficient. It is likely that this problem could be alleviated by having even shorter exposures in the input images (see sec.~\ref{section.data.capture}), which would better highlight subtle changes in sun intensity in those challenging cases. 
}

\section{Conclusion}
\label{section.conclusion}

In this work, we propose to learn the relationship between past and future photovoltaic power outputs using deep learning. In the context of short-term forecasting (also known as \emph{nowcasting}), our approach employs a \revise{deep LSTM-based network} which leverages past images as well as past \revise{photovoltaic} power output values, and outperforms simple baselines as well as more sophisticated neural network architectures based on multi-layer perceptrons. Our experiments demonstrate that three aspects are important in achieving this success: 1) modeling the temporal dynamics with the LSTM structure; 2) incorporating other sub-tasks in the learning process; and 3) exploiting sky images.

Despite its success, photovoltaic nowcasting is still a challenging task, for example, the constantly changing clouds are still quite hard to model and create inaccurate future power output predictions. It would be interesting to explore, in the future, how ``end-to-end'' deep learning techniques such as the ones we introduced can be combined with more explicit cloud movement modeling approaches, such as optical flow~\cite{dosovitskiy2015flownet}. Another future direction would be to implement this work on an actual system, and continuously learn to improve from the data that is captured live by a solar panel. In this context, a possible direction includes the exploration of active learning approaches which could, for example, better adapt to the properties of the solar panel used.

\section*{Acknowledgements}
The authors would like to posthumously thank the regretted Prof. Takashi Matsuyama, without whom this collaboration would not have been possible. We also thank Marie-Joëlle Gosselin for her help in running early experiments, and gratefully acknowledge the support of Nvidia with the donation of the GPUs used for this research. This work was supported by the NSERC Discovery Grant RGPIN2014-05314 and the FRQNT REPARTI Strategic Network.

\section*{References}
\bibliography{main.bib} 
\end{document}